\documentclass{article}

\usepackage{microtype}
\usepackage{graphicx}
\usepackage{subfigure}
\usepackage{booktabs} 

\usepackage{hyperref}

\usepackage[accepted]{icml2021}
\usepackage{amsfonts}

\icmltitlerunning{Promises and Pitfalls of Black-Box Concept Learning Models}

\begin{document}

\twocolumn[
\icmltitle{Promises and Pitfalls of Black-Box Concept Learning Models}



\icmlsetsymbol{equal}{*}

\begin{icmlauthorlist}
\icmlauthor{Anita Mahinpei}{equal,to}
\icmlauthor{Justin Clark}{equal,to}
\icmlauthor{Isaac Lage}{to}
\icmlauthor{Finale Doshi-Velez}{to}
\icmlauthor{Weiwei Pan}{to}
\end{icmlauthorlist}

\icmlaffiliation{to}{Harvard University}

\icmlcorrespondingauthor{Anita Mahinpei}{amahinpei@g.harvard.edu}
\icmlcorrespondingauthor{Justin Clark}{jclark@cyber.harvard.edu}

\icmlkeywords{Interpretable Machine Learning, Concept Learning, Concept Bottleneck Models, Concept Whitening Models}

\vskip 0.3in
]



\printAffiliationsAndNotice{\icmlEqualContribution} 

\begin{abstract}
Machine learning models that incorporate concept learning as an intermediate step in their decision making process can match the performance of black-box predictive models while retaining the ability to explain outcomes in human understandable terms. However, we demonstrate that the concept representations learned by these models encode information beyond the pre-defined concepts, and that natural mitigation strategies do not fully work, rendering the interpretation of the downstream prediction misleading. We describe the mechanism underlying the information leakage and suggest recourse for mitigating its effects.
\end{abstract}

\section{Introduction}
\label{intro}
When human decision makers need to  meaningfully interact with machine learning models, it is often helpful to explain the decision of the model in terms of intermediate concepts that are human interpretable. For example, if the raw data consists of high-frequency accelerometer measurements, we might want to frame a Parkinson's diagonsis in terms of a concept like ``increased tremors."

However, most deep models are trained end-to-end (from raw input to prediction) and, although there are a number of methods that perform post-hoc analysis of information captured by intermediate layers in neural networks (e.g. \citet{ghorbani2019towards, kim2018interpretability, zhou2018interpreting}), there is no guarantee that this information will naturally align with human concepts \citep{chen2020concept}. 

For this reason, a number of recent works propose explicitly aligning intermediate neural network model outputs with pre-defined expert concepts (e.g. increased tremors) in supervised training procedures (e.g. \citet{koh2020concept, chen2020concept, kumar2009attribute, lampert2009learning, de2018clinically, yi2018neural,bucher2018semantic,losch2019interpretability}). In each case, the neural network model learns to map raw input to concepts and then map those concepts to predictions. We call the mapping from input to concepts a Concept Learning Model (CLM), although this mapping may not always be trained independently from the downstream prediction task. Models that incorporate a CLM component have been shown to match the performance of complex black-box prediction models while retaining the interpretability of decisions based on human understandable concepts, since for these models, one can explain the model decision in terms of intermediate concepts.

Unfortunately, recent work noted that black-box CLMs do not learn as expected. Specifically, \citet{margeloiu2021concept} demonstrate that outputs of CLMs used in Concept Bottleneck Models (CBMs) encode more information than the concepts themselves. This renders interpretations of downstream models built on these CLMs unreliable (e.g. it becomes hard to isolate the individual influence of a concept like ``increase in tremors" if the concept representation contains additional information). The authors posit that outputs of CLMs are encouraged to encode additional information about the task label when trained jointly with the downstream task model. They suggest that task-blind training of the CLM mitigates information leakage. Alternatively, Concept Whitening (CW) models, which explicitly decorrelate concept representations during training, also have the potential to prevent information leakage \citep{chen2020concept}. 

In this paper, we observe that the issue of information leakage in concept representations is even more pervasive and consequential than indicated in exiting literature, and that existing approaches not completely address the problem.  We demonstrate that CLMs trained with natural mitigation strategies for information leakage suffer from it in the setting where concept representations are soft---that is, the intermediate node representing the concept is a real-valued quantity that corresponds to our confidence in the presence of the concept. Unfortunately, soft representations are used in most current work built on CLMs \citep{koh2020concept, chen2020concept}. Specifically, we (1) demonstrate how extra information can continue to be encoded in the outputs of CLMs and when exactly this leakage will occur; (2) demonstrate that mitigation techniques -- task-blind training, adding unsupervised concepts dimensions to account for additional task-relevant information, and concept whitening -- do not fully address the problem; and (3) suggest strategies to mitigate the effect of information leakage in CLMs.

\section{Background}

\textbf{The Concept Bottleneck Model} Consider training data of the form $\{(x_i, y_i, c_i)\}_{i=1}^{n}$ where $n$ is the number of observations, $x_i \in \mathbb{R}^d$ are inputs with $d$ features, $y_i \in \mathbb{R}$ are downstream task labels, and $c_i \in \mathbb{R}^k$ are vectors of $k$ pre-defined concepts. A Concept Bottleneck Model (CBM) \citep{koh2020concept} is the composition of a function, $g : X \rightarrow C$, mapping inputs $x$ to concepts $c$, and a function $h : C \rightarrow Y$, mapping concepts $c$ to labels $y$. We refer to $g$ as the CLM component. The functions $g$ and $h$ are parameterized by neural networks and can be trained independently, sequentially or jointly (details in Appendix Section \ref{sec:training_cbm}).

\textbf{Concept Whitening Model} In the Concept Whitening  (CW) model \citep{chen2020concept}, we similarly divide a neural network $f: X \rightarrow Y$ into two parts: 1) a feature extractor $g: X \rightarrow Z$ that maps the inputs to the latent space $Z$ and 2) and a classifier $h: Z \rightarrow Y$ that maps the latent space to the labels. We refer to $g$ as the CLM component. While the model is trained to predict the downstream task, each of the $k$ concepts of interest is aligned with one of the latent dimensions extracted by $g$: a pre-defined concept, $C$, is aligned to a specific latent dimension by applying a rotation to the output of $g$, such that the axis in the latent space along which pre-defined examples of $C$ obtain the largest value, is aligned with the latent dimension chosen to represent $C$. The latent dimensions are decorrelated to encourage independence of concept representations. Any extra latent dimensions are left unaligned but still go through the decorrelation process.

\section{Extraneous Information Leakage in Soft Concept Representations}
\label{sec:pit-falls}

When the concept representation is soft, as in most current work using CLMs (\citet{koh2020concept}, \citet{chen2020concept}), these values not only encode for the pre-defined (that is, the desired) concepts, but they also \emph{unavoidably} encode the distribution of the data in the feature space. This allows non-concept-related information to leak through, rendering flawed interpretations of the prediction based on the concepts.

Previous work observed that this information leakage happens when the CLM is trained jointly with the downstream task model \cite{margeloiu2021concept}: in this case, the joint training objective encourages the concept representations to encode for all possible task-relevant information.

Below, we demonstrate information leakage occurs \emph{even without joint training}, and that natural mitigation strategies are not sufficient to solve it.  Specifically, we consider three methods for mitigating information leakage (1) sequential training of CLMs (where the CLM is trained prior to consideration of any downstream tasks), (2) adding unsupervised concepts dimensions to account for extra task-relevant information, and (3) decorrelating concept representations through concept whitening.  We explain why none succeed in completely preventing leakage.

\textbf{Leakage Occurs When the CLM is Trained Sequentially} When the CLM $g : X \rightarrow C$ is trained separately from the concept-to-task model $h : C \rightarrow Y$ rather than jointly, one might expect no information leakage; the task-blind training of the CLM should only encourage the model to capture the concepts and not any additional task-relevant information. We show that, surprisingly, the soft concept representations of these CLMs nonetheless encode much more information than the concepts themselves, even when the concepts are irrelevant for the downstream task. 

Consider the task of predicting the parity of digits in a subset of MNIST, in which there are equal numbers of odd and even digits. Define two binary concepts: whether the digit is a four and whether the digit is a five. We consider an extreme case where there are zero instances of fours or fives in both training and hold-out data. In this case, the concepts are clearly task-irrelevant and a classifier for parity built on these concept labels should do no better than random guessing -- we'd expect a predictive accuracy of 50\%. 

We set up a CBM as follows: we parameterize the CLM, i.e. the feature-to-concept model ($g(x) \mapsto c$), as a feed-forward network (two hidden layers, 128 nodes each, ReLU activation) whose output is passed through a sigmoid; we parameterize the concept-to-task model with a neural network with the same architecture. Using the sequential training approach, we first train the CLM. 
Fixing the trained CLM, we train the concept-to-task model to predict parity based on the concept probabilities output by the CLM. The latter model achieves a test accuracy of 69\%, far higher than the expected 50\%! 

\begin{figure}[h!]
    \centering
    \includegraphics[scale=0.5]{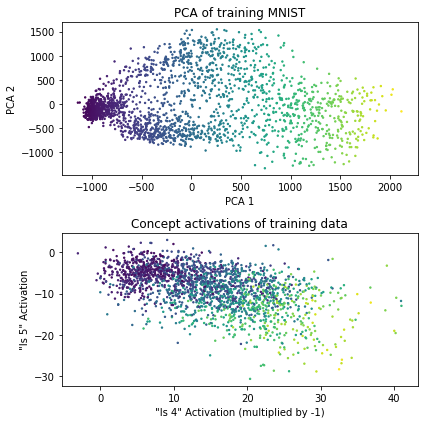}
    \caption{\textbf{Top}: The top two PCA dimensions of a curated subset of MNIST. \textbf{Bottom}: Activations for the two concepts of ``is 4" and ``is 5" for each observation. Colors in both indicate each observation's location along the first PCA dimension. Note that the ``is 4" concept preserves the ordering of the colors, and thus encoding the first PCA dimension (Pearson correlation of -0.72). The ``is 5" concept also helps preserve the first PCA dimension (Pearson correlation of -0.67). Together these soft concept representations encode the data distribution to allow for good task performance though both concepts are independent of the task.}
    \label{fig:mnist_activations}
\end{figure}

Why are the outputs from our CLM far more useful than the true concept labels for the downstream task?  The reason lies in the fact that concept classification probabilities encode much more information than whether or not a digit is a ``4" or a ``5."  The concept classification probability is a function of the distance from the input to the decision surface corresponding to that concept. These soft concept representations, therefore, \emph{necessarily} 
encode the distribution of the data along axes perpendicular to decision surfaces.  The downstream classifier $h: C \rightarrow Y$ takes advantage of this additional information to make its predictions.

We see this in the top panel of Figure \ref{fig:mnist_activations}, where we visualize the first two PCA dimensions of our MNIST dataset (the color indicates each observation's location along the first PCA dimension). In the bottom panel, we visualize the activation values of the output nodes corresponding to the concepts of ``4" and ``5". We see that the activations for the concept ``4'' roughly preserve the first PCA dimension of the MNIST data (``5" encodes similar information). In fact, predicting the task label using only the top PCA component of the data yields an impressive accuracy of 86\%.

In this example, the concepts ``4" and ``5" are semantically related to the downstream task, so is this why their soft representations encode salient aspects of the data?  Unfortunately, in Appendix Section \ref{sec:demo0}, we show that, even \emph{random} concepts (i.e. concepts defined by random hyperplanes in the input space with no real-life meaning) can capture much of the data distribution through soft representations. In fact, as the number of random concepts grow so does the degree of information leakage (as measured by the difference between the utility of hard and soft concept representations for the downstream prediction). This has significant consequences for decision-making based on interpretations of Concept Bottleneck Models. In particular, one cannot naively interpret the utility of the learned concepts representations as evidence for the correlation between the human concepts and the downstream task label - e.g. the fact that the \emph{predicted probability} of an increase in tremors is highly indicative of Parkinson's disease does not imply that there is significant correlation between an increase in tremors and Parkinson's in the data.

Finally, although all soft concept representations encode more than we might desire, in Appendix Section \ref{sec:demo1}, we show that the extent to which information about the data distribution leaks into the soft concept representations - and correspondingly, the extent to which soft outputs from CLMs become more useful than true concept labels for downstream tasks - is sensitive to modeling choices (hyperparameters, architecture and training). 

\textbf{Leakage Occurs When the CLM is Trained with Added Capacity to Represent Latent Task-Relevant Dimensions} Another approach for getting more pure concepts is to train a CLM with additional unsupervised concept dimensions that can be used to capture task-relevant concepts not included in the set of concept labels (e.g. \citet{chen2020concept}). The hope here is that, should the original, curated concepts be insufficient for the downstream prediction, these additional dimensions will capture what is necessary and leave the original concept dimensions interpretable.  
Note that, in this case, the CLM cannot be trained sequentially since we do not have labels for the missing concepts, so optimization must be done jointly.

Unfortunately, leakage still occurs with this approach.
Concepts, both pre-defined and latent, continue to be entangled in the learned representations, even when they are fully independent in the data. To demonstrate that leaving extra space for unlabeled concepts in the representation does not solve the leakage issue, we generate a toy dataset with 3 
 concepts, independent from each other, that are used to generate the label. We generate the dataset with seven input features, $\{\left(f_1(x_n, y_n, z_n), \ldots, f_7(x_n, y_n, z_n)\right)\}$, where $X$, $Y$, and $Z \sim \mathcal{N}(0, 4)$ and each $f_i$ is a non-invertible, nonlinear function (details in Appendix Section \ref{sec:demo2}). For concepts, we define three binary concept variables $x_{+}$, $y_{+}$, and $z_{+}$, each indicating whether the corresponding value of $x_n, y_n, z_n$ is positive. The downstream task is to identify whether at least two of the three values $x_n, y_n, z_n$ are positive (i.e. whether the sum of $x_{+}$, $y_{+}$, and $z_{+}$ is greater than 1). 

We consider three models: (M1) a CBM with a complete concept set (i.e. 3 bottleneck nodes, aligned to $x_{+}$, $y_{+}$, and $z_{+}$ respectively) as a baseline where there are no missing concepts and the model has sufficient capacity; (M2) a CBM with an incomplete concept set (i.e. 2 bottleneck nodes aligned to $x_{+}$ and $y_{+}$ respectively) and insufficient capacity (no additional node for $z_{+}$); and (M3) a CBM with an incomplete concept set, 2 bottleneck nodes aligned to $x_{+}$ and $y_{+}$, and one unaligned bottleneck node representing the latent concept $z_{+}$, as the main case of interest where we know only some of the concepts but leave sufficient capacity for additional important concepts. All models are jointly trained to ensure that unaligned bottleneck nodes can capture the task-relevant latent concept $z_+$.

We find that all 3 models are predictive in the downstream task, despite M2 having insufficient capacity for the relevant concepts and no supervision for $z_{+}$.  M2 is able to achieve an AUC of $0.999$ on the downstream task, whereas a standard neural network trained to predict the task labels using the ground-truth hard labels $x_{+}$ and $y_{+}$ obtains an AUC of $0.875$. 
The exceptional performance of M2 suggests that joint training encourages the soft representations of $x_+$ and $y_+$ to encode additional information necessary for the downstream task. 

To determine how these representations are encoding information about the latent concept $z_+$ in these settings, we test the concept purity. Specifically, we measure whether we can predict concept labels based on the soft output of each individual concept node by reporting their AUC scores.  If the concept is predictable from the node aligned with it, but not from the other nodes aligned with the other 2 concepts, then we consider the concept node \emph{pure} (note that the 3 concepts are mutually independent by construction). For a node that represents its aligned concept \emph{purely}, we expect an AUC close to 1.0 when predicting the concept from that node, and an AUC close to 0.5 (random guessing) when predicting it from the other 2 nodes.

For all models (even when the concept dimensions are supervised by the complete concept set during training), we observe impurities in all bottleneck dimensions. Although the aligned bottleneck dimensions are most predictive of the concepts with which they are aligned, they also have AUC's greater than 0.6 for concepts with which they are not aligned (Appendix Tables  \ref{auc-complete}, \ref{auc-incomplete}, \ref{auc-latent}). This supports our claims above that soft representations may entangle concepts, even when labeled; this also supports claims from \citet{margeloiu2021concept} and \citet{koh2020concept} about joint training potentially causing additional concept entanglement.

Having an incomplete concept set exacerbates the leakage problem as the model is forced to relay a lot more information about the missing concept through the bottleneck dimensions. For M2, the concept dimensions aligned to $x_+$ and $y_+$ each predicts $z_+$ with AUC of approximately 0.75.
Unfortunately, adding bottleneck dimensions in an attempt to capture the missing concept does not prevent leakage. 
We notice that the added bottleneck dimension in M3 does not consistently align with the missing concept across random trials (this dimension predicts $z_+$ with an AUC of $0.4 \pm 0.4$). Furthermore, the unaligned dimension sometimes contains information about $x_+$, $y_+$, further compromising the interpretability of the model (details in Appendix Section \ref{sec:demo2}).

In all 3 models, soft representations entangle concepts, rendering interpretations of the CBM potentially misleading. For example, in these experiments, we do not recover feature importance
of the ground truth concept-to-task model when using entangled concept representations for our downstream task model -- according to the learned concept-to-task model, the concept $x_+$ appears to be more important for the downstream prediction than in the ground truth model.

\textbf{Leakage Occurs When the Concept Representations are Decorrelated}
We just observed that joint training results in impure concepts, even when extra capacity is given to learn additional task-relevant concept dimensions.  A natural solution might be to encourage independence among concept representations during training. The CW model implements a form of this training---it decorrelates the concept representations. However, we find that even in these decorrelated representations, information leakage occurs. We demonstrate that concepts can be predicted from other (even aligned) latent dimensions after concept whitening, which can potentially confound interpretation of predictions based on the concept representation.

To demonstrate that this information leakage can occur between concepts after concept whitening, we consider the task of classifying whether MNIST digits are less than 4. We create the following three binary concepts: 1) ``is the number 1''  2) ``is the number 2''  3) ``is the number 3''. We train a CW model, aligning two latent dimensions to the first two concepts, and leaving a number of latent dimensions unaligned to capture the missing concept.

We find that although the learned concept representations satisfy all three purity criteria described in \citet{chen2020concept} -- in particular, the representations are decorrelated -- we can still predict any of the three concepts from any of the latent concept dimensions (both aligned and unaligned). This is because 1) purity is computed based on single activation values that summarize high-dimensional concept representations; thus while these single summary values are predictive of only one concept, the high-dimensional concept representation (used by the downstream task model) can encode information about other concepts; and 2) decorrelating concept representations does not remove all statistical dependence; two representations can still share high mutual information while being uncorrelated. Again, importantly, we do not recover the ground truth feature importance of the three concepts when interpreting the downstream task model built on whitened concept representations. Details can be found in Appendix Section \ref{sec:demo3}.

\section{Avoiding the Pit-Falls of CLMs}

In the following, we outline ways to mitigate the negative effects of information leakage.

\textbf{Soft Versus Hard Concept Representations} Since all of the pathologies that we observe arise from the flexibility of soft concept representations, one might be tempted to propose always using hard concept representations (one-hot encodings). 
However, in Appendix Section \ref{sec:demo0}, we observe that (when the data manifold is low-dimensional, as in MNIST) even a modest number of semantically meaningless random hard concepts can capture more information about the data distribution than we might expect. This indicates that information leakage may always be an issue with black-box CLMs, regardless of the form of concept representation. 

 Furthermore, we argue that the analysis of soft concept representations yields important insights for model improvement/debugging. Specifically, if the downstream label is better predicted with soft concept representations than with hard, it may indicate that the concept set is not relevant for the task (as in the case of the concepts of ``4" and ``5" when predicting the parity of digits in a dataset without 4's and 5's), and a new set of concepts must be sought. Alternatively, this difference in utility may indicate that the concepts should be modified, with the help of human experts. In Appendix Section \ref{sec:demo4}, we describe a toy example in which the pre-defined concept set is related to but not strongly predictive of the downstream label, where domain expertise allows us to refine them into useful sub-concepts. In fact, we are currently exploring strategies to leverage domain expertise to refine or suggest new concepts, based on learned soft concept representations, that are  human-interpretable and predictive of the downstream task.

\textbf{Disentangling Concepts}
While CW models encourage concept dimensions to be uncorrelated, this does not completely prevent information leakage -- we show that each concept dimension can still encode for multiple concepts and that the concept dimensions can nonetheless be statistically dependent. Thus, we argue that CLM training should explicitly minimize mutual information between concept dimensions -- both aligned and unaligned -- as in \citet{klys2018learning}, if we believe the concepts to be independent. However, we note that if there are multiple statistically independent latent concepts, it is still possible for each concept dimension to encode for multiple concepts. Thus, we again advocate for domain expert supervision in the definition and training of the CLM, bringing more transparency to the relationship between input dimensions and learned concepts.

\section{Conclusion}
In this paper, we analyze pitfalls of black-box CLMs stemming from the fact that soft concept representations learned by these models encode undesirable additional information. We highlight scenarios wherein this information leakage negatively impacts the interpretability of the downstream predictive model, as well as describe important insights that can be gained from understanding the additional information contained in soft concept representations. 

\section{Acknowledgments} WP was funded by the Harvard Institute of Applied Computation Science. IL was funded by NSF GRFP (grant no. DGE1745303).  FDV was supported by NSF CAREER 1750358.

\bibliography{main}

\begin{thebibliography}{13}
\providecommand{\natexlab}[1]{#1}
\providecommand{\url}[1]{\texttt{#1}}
\expandafter\ifx\csname urlstyle\endcsname\relax
  \providecommand{\doi}[1]{doi: #1}\else
  \providecommand{\doi}{doi: \begingroup \urlstyle{rm}\Url}\fi

\bibitem[Bucher et~al.(2018)Bucher, Herbin, and Jurie]{bucher2018semantic}
Bucher, M., Herbin, S., and Jurie, F.
\newblock Semantic bottleneck for computer vision tasks.
\newblock In \emph{Asian Conference on Computer Vision}, pp.\  695--712.
  Springer, 2018.

\bibitem[Chen et~al.(2020)Chen, Bei, and Rudin]{chen2020concept}
Chen, Z., Bei, Y., and Rudin, C.
\newblock Concept whitening for interpretable image recognition.
\newblock \emph{Nature Machine Intelligence}, 2\penalty0 (12):\penalty0
  772--782, 2020.

\bibitem[De~Fauw et~al.(2018)De~Fauw, Ledsam, Romera-Paredes, Nikolov, Tomasev,
  Blackwell, Askham, Glorot, O’Donoghue, Visentin, et~al.]{de2018clinically}
De~Fauw, J., Ledsam, J.~R., Romera-Paredes, B., Nikolov, S., Tomasev, N.,
  Blackwell, S., Askham, H., Glorot, X., O’Donoghue, B., Visentin, D., et~al.
\newblock Clinically applicable deep learning for diagnosis and referral in
  retinal disease.
\newblock \emph{Nature medicine}, 24\penalty0 (9):\penalty0 1342--1350, 2018.

\bibitem[Ghorbani et~al.(2019)Ghorbani, Wexler, Zou, and
  Kim]{ghorbani2019towards}
Ghorbani, A., Wexler, J., Zou, J., and Kim, B.
\newblock Towards automatic concept-based explanations.
\newblock \emph{arXiv preprint arXiv:1902.03129}, 2019.

\bibitem[Kim et~al.(2018)Kim, Wattenberg, Gilmer, Cai, Wexler, Viegas,
  et~al.]{kim2018interpretability}
Kim, B., Wattenberg, M., Gilmer, J., Cai, C., Wexler, J., Viegas, F., et~al.
\newblock Interpretability beyond feature attribution: Quantitative testing
  with concept activation vectors (tcav).
\newblock In \emph{International conference on machine learning}, pp.\
  2668--2677. PMLR, 2018.

\bibitem[Klys et~al.(2018)Klys, Snell, and Zemel]{klys2018learning}
Klys, J., Snell, J., and Zemel, R.
\newblock Learning latent subspaces in variational autoencoders.
\newblock \emph{arXiv preprint arXiv:1812.06190}, 2018.

\bibitem[Koh et~al.(2020)Koh, Nguyen, Tang, Mussmann, Pierson, Kim, and
  Liang]{koh2020concept}
Koh, P.~W., Nguyen, T., Tang, Y.~S., Mussmann, S., Pierson, E., Kim, B., and
  Liang, P.
\newblock Concept bottleneck models.
\newblock In \emph{International Conference on Machine Learning}, pp.\
  5338--5348. PMLR, 2020.

\bibitem[Kumar et~al.(2009)Kumar, Berg, Belhumeur, and
  Nayar]{kumar2009attribute}
Kumar, N., Berg, A.~C., Belhumeur, P.~N., and Nayar, S.~K.
\newblock Attribute and simile classifiers for face verification.
\newblock In \emph{2009 IEEE 12th international conference on computer vision},
  pp.\  365--372. IEEE, 2009.

\bibitem[Lampert et~al.(2009)Lampert, Nickisch, and
  Harmeling]{lampert2009learning}
Lampert, C.~H., Nickisch, H., and Harmeling, S.
\newblock Learning to detect unseen object classes by between-class attribute
  transfer.
\newblock In \emph{2009 IEEE Conference on Computer Vision and Pattern
  Recognition}, pp.\  951--958. IEEE, 2009.

\bibitem[Losch et~al.(2019)Losch, Fritz, and
  Schiele]{losch2019interpretability}
Losch, M., Fritz, M., and Schiele, B.
\newblock Interpretability beyond classification output: Semantic bottleneck
  networks.
\newblock \emph{arXiv preprint arXiv:1907.10882}, 2019.

\bibitem[Margeloiu et~al.(2021)Margeloiu, Ashman, Bhatt, Chen, Jamnik, and
  Weller]{margeloiu2021concept}
Margeloiu, A., Ashman, M., Bhatt, U., Chen, Y., Jamnik, M., and Weller, A.
\newblock Do concept bottleneck models learn as intended?
\newblock \emph{arXiv preprint arXiv:2105.04289}, 2021.

\bibitem[Yi et~al.(2018)Yi, Wu, Gan, Torralba, Kohli, and
  Tenenbaum]{yi2018neural}
Yi, K., Wu, J., Gan, C., Torralba, A., Kohli, P., and Tenenbaum, J.~B.
\newblock Neural-symbolic vqa: Disentangling reasoning from vision and language
  understanding.
\newblock \emph{arXiv preprint arXiv:1810.02338}, 2018.

\bibitem[Zhou et~al.(2018)Zhou, Bau, Oliva, and Torralba]{zhou2018interpreting}
Zhou, B., Bau, D., Oliva, A., and Torralba, A.
\newblock Interpreting deep visual representations via network dissection.
\newblock \emph{IEEE transactions on pattern analysis and machine
  intelligence}, 41\penalty0 (9):\penalty0 2131--2145, 2018.

\end{thebibliography}
\bibliographystyle{icml2021}

\appendix

\section{Training of Concept Bottleneck Models}
\label{sec:training_cbm}
There are three approaches for training a CBM, consisting of a function $g : X \rightarrow C$, mapping inputs $x$ to concepts $c$, and a function $h : C \rightarrow Y$, mapping concepts $c$ to labels $y$:
\begin{enumerate}
    \item Independently learning $g$ and $h$ by minimizing their respective loss functions $\sum_{i=1}^{n} \mathcal{L}_g(g(x_i); c_i)$ and $\sum_{i=1}^{n} \mathcal{L}_h(h(c_i); y_i)$.
    
    \item Sequentially learning $g$ and $h$ by first minimizing the loss function $\sum_{i=1}^{n} \mathcal{L}_g(g(x_i); c_i)$ and then minimizing the loss function $\sum_{i=1}^{n} \mathcal{L}_h(h(g(x_i)); y_i)$ using the outputs of $g$ as inputs to $h$.
    
    \item Jointly learning $g$ and $h$ by minimizing $\sum_{i=1}^{n}[ \mathcal{L}_h(h(g(x_i)); y_i) + \lambda \mathcal{L}_g(g(x_i); c_i)]$ where $\lambda$ is a hyperparameter that determines the tradeoff between learning the concepts and the labels.
\end{enumerate}

\section{Demonstration 1: Concept Representations Encode Data Distributions}\label{sec:demo1}

\begin{figure}[h!]
    \centering
    \includegraphics[scale=0.5]{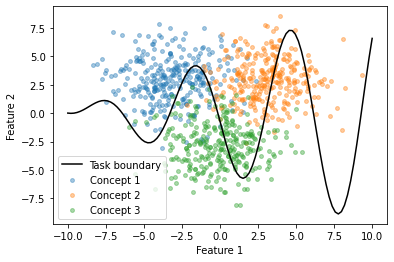}
    \caption{A synthetic dataset of two features, three concepts, and a somewhat complex task boundary that cuts across concepts.}
    \label{fig:features_concepts_task}
\end{figure}

\begin{figure}[h!]
    \centering
    \includegraphics[scale=0.5]{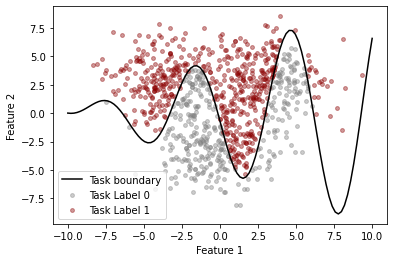}
    \caption{The same dataset as Figure \ref{fig:features_concepts_task} but with colors indicating task labels. This is the dataset prior to transformations present in Figures \ref{fig:concept_preds_to_task} and \ref{fig:sigmoided}.}
    \label{fig:features_task}
\end{figure}

We consider a simple synthetic binary classification task where the data consist of two features and three mutually exclusive, binary concepts (see Figures \ref{fig:features_concepts_task} and \ref{fig:features_task}). There are 300 observations for each concept label, and approximately 60\% of the observations have a positive task label.

We build models for four tasks:
\begin{itemize}
    \item Predict the task labels from the features, $h(x) \to \hat{y}$
    \item Predict the task labels from the concepts, $f(c) \to \hat{y}$
    \item Predict the concepts from the features, $g(x) \to \hat{c}$
    \item Predict the task labels from the predicted concepts, $f(g(x)) \to \hat{y}$
\end{itemize}
For each model, we use a simple MLP with three hidden layers of 32 nodes each and ReLU activations. When predicting the task label, the final layer has a sigmoid activation, and we use a classification threshold of 0.5.

We want to know how predictive our raw concept labels are of the task labels. We train a model with an architecture similar to above. These raw concept labels consist of only the values $[0, 0, 1]$, $[0, 1, 0]$, and $[1, 0, 0]$, representing each of the mutually exclusive concepts. The model learns that two concepts are associated with the positive task label (the blue and orange concepts in Figure \ref{fig:features_concepts_task}) and the remaining concept is associated with the negative task label, which achieves an accuracy of 74.5\% on a test set. This is significantly better than the 60\% we would expect of random guessing, so the concepts are not independent of the task.

We now move to building the sequential concept model. The first component, $g(x) \to \hat{c}$, predicts the concept labels from the features. The model architecture is the same as above, except the output layer contains three nodes with linear activations, and we use mean-squared error loss. (We chose linear activations because they make the behavior we're trying to show more easily visible geometrically, but the same behavior exists with softmax or sigmoid activations.) If we take the maximum activation of the output nodes as our concept prediction, this model achieves 87.0\% accuracy on a test set.

We now train the second component to complete our sequential concept model, $f(\hat{c}) \to \hat{y} = f(g(x)) \to \hat{y}$. We know our concepts can predict our task labels with 74.5\% accuracy and our features can predict our concepts with 87.0\% accuracy. If we train the second component of our concept model to convergence, it achieves an accuracy of 95.9\% on the downstream task. On first blush, this is surprising as it is significantly higher than the performance of the model the predicts the task from the concept labels.

\begin{figure}[h!]
    \centering
    \includegraphics[scale=0.4]{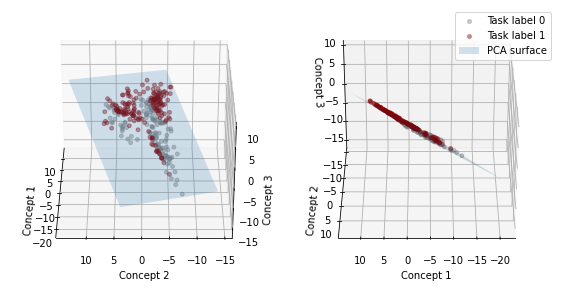}
    \caption{Concept activations output from the features-to-concepts model. Because the three concepts are mutually exclusive, only two concept activations are independent. Hence, the three concept activations can be projected onto a plane without loss of information. \textbf{Left}: One angle showing the dataset. \textbf{Right}: A second angle showing the data lying on a plane.}
    \label{fig:planar_activations}
\end{figure}

\begin{figure}[h!]
    \centering
    \includegraphics[scale=0.5]{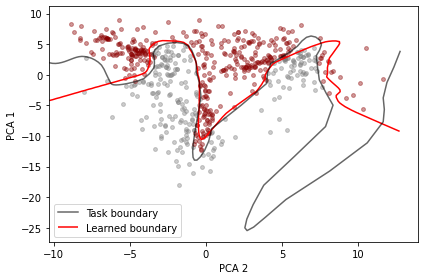}
    \caption{The first two PCA dimensions of the concept activations, along with the synthetic task boundary and learned task boundary in the same PCA space. The task boundary and dataset are only modestly transformed from the original dataset seen in Figure \ref{fig:features_task}, and the learned task boundary appears similar to that learned directly from the features (Figure \ref{fig:features_to_task}). The concept predictions $\to$ task model achieves 95.9\% accuracy.}
    \label{fig:concept_preds_to_task}
\end{figure}

\begin{figure}
    \centering
    \includegraphics[scale=0.5]{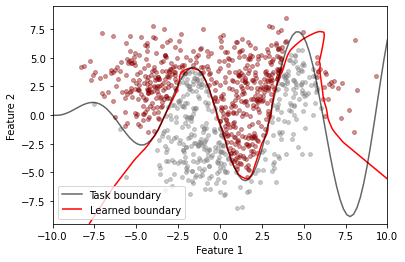}
    \caption{The dataset and learned task boundary if the task is learned directly from the features. The features $\to$ task model achieves 99.3\% accuracy.}
    \label{fig:features_to_task}
\end{figure}

To understand what is happening, we can visualize the three-dimensional concept activations of the first component of the concept model, $g(x) \to \hat{c}$ (Figure \ref{fig:planar_activations}). As the concepts are mutually exclusive, there are only two linearly independent concept vectors, so we can accurately summarize our three-dimensional concept activations with the first two dimensions of a PCA decomposition. We can also pass our original synthetic decision boundary through the first component of our sequential concept model, $g(x)$, and through our PCA transformation to see how it is affected. We can see the resulting graph in Figure \ref{fig:concept_preds_to_task}. If we train a model to predict the task labels from the features directly, with no concepts ($h(x) \to \hat{y}$), we achieve a test accuracy of 99.3\% (Figure \ref{fig:features_to_task}). Comparing Figure \ref{fig:concept_preds_to_task} to Figure \ref{fig:features_to_task}, we can see that $g(x)$ actually did very little; the geometry of the features and task boundary are mostly intact. $g(x)$ may have aligned concept labels with specific neurons in the overall concept model, but it is essentially a mild transformation of the features that resulted in a slightly harder task. When adding concepts to a model, we should not interpret the minor drop in performance from 99.3\% to 95.9\% as signifying a relationship between the concepts and the task -- we should instead interpret it as signifying a relationship between the features, the concepts, and modeling decisions made when predicting the concepts from the features.

\begin{figure}
    \centering
    \includegraphics[scale=0.5]{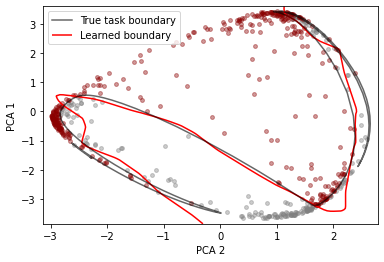}
    \caption{The first two PCA dimensions of the sigmoided concept activations, along with the synthetic task boundary and learned task boundary in the same PCA space. The synthetic task boundary is severly transformed, leading to a harder optimization problem. The concept predictions $\to$ task model with sigmoid activations achieves 89.5\% accuracy.}
    \label{fig:sigmoided}
\end{figure}

Modeling decisions are involved because the success that $f(\hat{c})$ has in predicting the task labels depends on the complexity of the task decision boundary after passing through $g(x)$. This complexity depends not only on the data, but on the architecture, activation functions, and learning rate used when training $g(x) \to \hat{c}$. For example, in $g(x)$ above, the activations in the output layer were identity functions. If we instead use sigmoid activations to ensure our concept predictions lie between 0 and 1 (see Figure \ref{fig:sigmoided}), the feature data is transformed in a more extreme way which results in task performance dropping to 89.5\%. If the downstream task performance of a set of concepts is dependent on architectural choices (e.g. particular activation functions), interpreting that performance as a property of the relationship between the concept set and the task is a mistake.

This propagation of feature information is the same mechanism that explains the performance of the concept model in the MNIST example. While there is no additional information provided by the concepts, they allow enough of the original feature information to pass through $g(x)$ that $f(g(x))$ can perform decently.

\section{Demonstration 2: Sufficiently Large Number of Random Concepts Can Encode Data Distribution via Soft Representations}
\label{sec:demo0}

We subset both the training and test splits of MNIST to include only 500 images each of the digits 0, 1, 6, and 7. The task is to predict whether or not each image depicts an even digit. We perform this task using two models trained sequentially: one that predicts concept labels from features, $g(x) \to \hat{c}$, and one that predicts the task label from those concept predictions, $f(\hat{c}) \to \hat{y}$. We refer to these concept predictions, $\hat{c}$, as the ``soft" representation of the concept in contrast to the ``hard", binary representation of the concept. Each model is a fully connected feed-forward neural network with two hidden layers of 128 nodes each, ReLU activations and binary cross entropy loss.

Instead of using human interpretable concepts, we generate random concept labels. We do this by generating random hyperplanes in the feature space and labeling all images that fall on one side of these hyperplanes as ``true" examples while the remaining are ``false" examples. The equation for a hyperplane in this context can be given as $a_1x_{i,1} + a_2x_{i,2} + ... + a_kx_{i,k} = b$ where $i$ is the index of the given observation, $k$ is the dimension of the feature space, $a_k$ is an arbitrary coefficient, $x_{i,k}$ is the value of the $k$th feature of the $i$th observation, and $b$ is an arbitrary number. To generate random hyperplanes, we first generate coefficients $a_1...a_k$ by randomly sampling from the uniform distribution from 0 to 1. For each observation in the training set, we calculate the dot product of these random coefficients and the observation's feature values, $a_1x_{i,1} + ... + a_kx_{i,k} = s_i$. This gives us one number, $s_i$, for each training image. We then randomly sample from the uniform distribution bounded by the minimum and maximum of this set of dot products, $b \sim \mathrm{Unif}(\min([s_1,...,s_n]), \max(s_1,...,s_n)])$, where $n$ is the total number of observations. Sampling $b$ in this way ensures the generated hyperplane passes through the training dataset. Observations in both the training and test sets were then assigned concept values based on the inequality $a_1x_{i,1} + ... + a_kx_{i,k} < b$. This process was repeated for as many random concepts we chose to generate.

We fit models with an increasing number of random hyperplane concepts. For each number of concepts, we first trained $g(x) \to \hat{c}$ to generate the ``soft" concept representations $\hat{c}$ and then trained $f(\hat{c}) \to \hat{y}$. Hyperparameters for each number of concepts were tuned to ensure the models trained to convergence. For 10 runs with different seeds, we recorded the accuracy of the composition of these models, $f(g(x)) \to \hat{y}$, on the test set.

As a point of comparison, we also trained models to predict each digit's parity from the hard representation of each of set of random hyperplane concepts using the same architecture, but with different hyperparameters to ensure convergence.

\begin{figure*}
    \centering
    \includegraphics[scale=0.3]{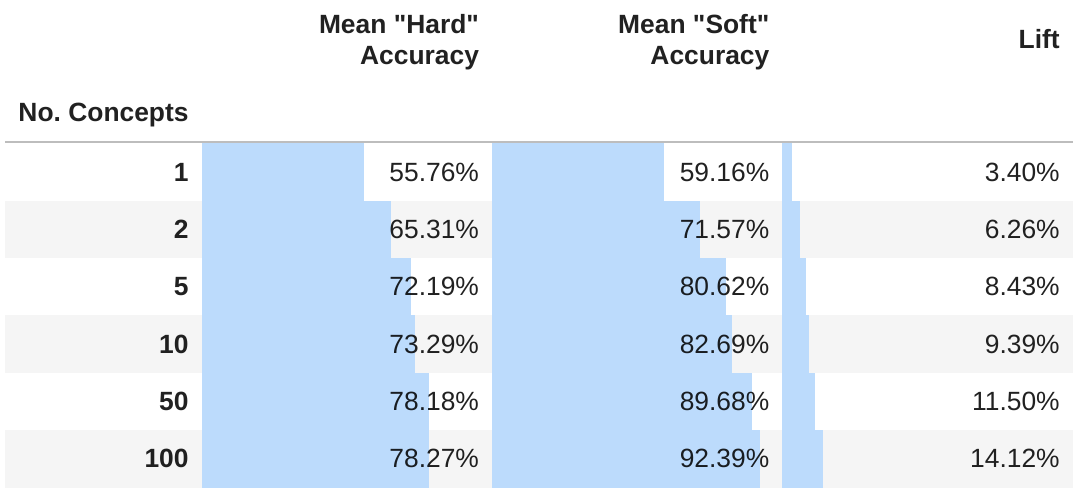}
    \caption{Performance of ``hard" and ``soft" random concept representations on the task of predicting MNIST digit parity. Accuracies are averaged over 10 runs. Performance increases but starts to diminish as more hard concepts are introduced. The soft representations of the same concepts have better performance, and the performance doesn't diminish in the same way. We suggest this is because more random concepts allow for more feature information to pass through.}
    \label{fig:increasing_random_concepts}
\end{figure*}

Figure \ref{fig:increasing_random_concepts} shows the results of this experiment. As more random concepts are added, the predictiveness of the hard concept representations increases, notably indicating that \emph{even hard concepts representations can encode more information than we intend}. The predictiveness of the soft concept representations increases even more so -- these soft representations always perform better than the hard representations when used in the downstream task. When predicting the task labels directly from the 784 pixel values, a model of similar architecture achieves 99\% accuracy on the test set. We can see that as the number of random features increases, the models approach this performance, thus showing that most of the feature information has passed through the concept modeling process.

\section{Demonstration 3: Representations Entangling Concepts}
\label{sec:demo2}

A normal distribution $\mathcal{N}(0, 2)$ is used to independently generate three coordinates $x$, $y$, and $z$ for our toy dataset. The input features to our model are defined by a set of seven non-invertible function transformations, six of which are created by applying $\beta(w) = \sin(w) + w$ and  $\alpha(w) = \cos(w) + w$ to each of our three coordinates. The final feature  is defined as the sum of squares of the coordinates. The concepts are the binary variables $x_{+}$, $y_{+}$, and $z_{+}$, each indicate whether the corresponding values $x_n, y_n, z_n$ have positive values. The downstream task is to identify whether at least two of the values $x_n, y_n, z_n$ are positive, that is, whether the sum of concepts $x_{+}$, $y_{+}$, and $z_{+}$ is greater than 1. Our training dataset contains 2000 examples while our test dataset contains 1000 examples.

We first set up a standard neural network model with four layers (8, 6, 4, 1 nodes) to predict labels from features. The model achieves an AUC of $0.99$. We then create a jointly trained CBM by setting the bottleneck at the second layer of our standard model. We train the CBM both with the complete concept set (ie. 3 bottleneck nodes) and the incomplete concept set that only uses concepts $x_{+}$ and $y_{+}$ (ie. 2 bottleneck nodes). A naive approach to address the incompleteness of the concept set could be to include latent dimensions in the bottleneck that are not aligned with any concepts and see whether they will automatically align with the missing concepts during training. Thus, we also build a CBM that is trained with the incomplete concept set including $x_{+}$ and $y_{+}$ but with three bottleneck nodes; the first two nodes will be aligned with $x_{+}$ and $y_{+}$ while the third node will be the latent dimension. We train the CBMs for 350 epochs using the Adam optimizer with a learning rate of 0.001. The $\lambda$ hyperparameter from the jointly trained CBM loss function is set to  $0.001$, $0.01$, $0.1$, $1.0$, and $10.0$. $\lambda=0.1$ generally provides the best trade-off between concept prediction accuracy and downstream task accuracy. 

\begin{figure}[t]
    \centering
    \includegraphics[scale=0.5]{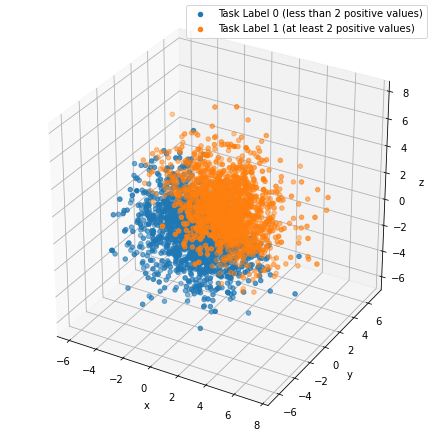}
    \caption{The distribution of the coordinates $X, Y, Z \sim \mathcal{N}(0, 4)$ generated for demonstration 3. Data points have a downstream task label of 1, if at least two of their coordinate values are positive.}
    \label{fig:incompleteness_cbm_data}
\end{figure}

Despite having no latent dimensions and an incomplete set of concepts, the second CBM model is able to achieve high downstream task AUCs that are greater than $0.98$ and accuracies that are greater than $90\%$ for all $\lambda$ less than or equal to $0.1$. We train a standard neural network with two layers (4 and 1 nodes) using the original concepts $x_{+}$ and $y_{+}$ to predict the task labels and obtain an AUC of $0.875$ and an accuracy of $75\%$. This makes sense; algebraically, we have $\frac{2}{3}$ of the information needed for the downstream task, and at best, we can correctly guess the third missing concept half of the time (since the binary concepts are balanced). This brings the maximum possible accuracy to about $83\%$, yet our CBM achieved much greater accuracies. These results bring into question the interpretability of CBMs when our concept set is incomplete.  When $\lambda$ is small, the model seems to relay information through the bottleneck beyond what is provided in our known concepts thus compromising model interpretability. When $\lambda$ is large, the model will show a significant drop in accuracy compared to a standard neural network model.

\begin{figure*}
    \centering
    \includegraphics[scale=0.4]{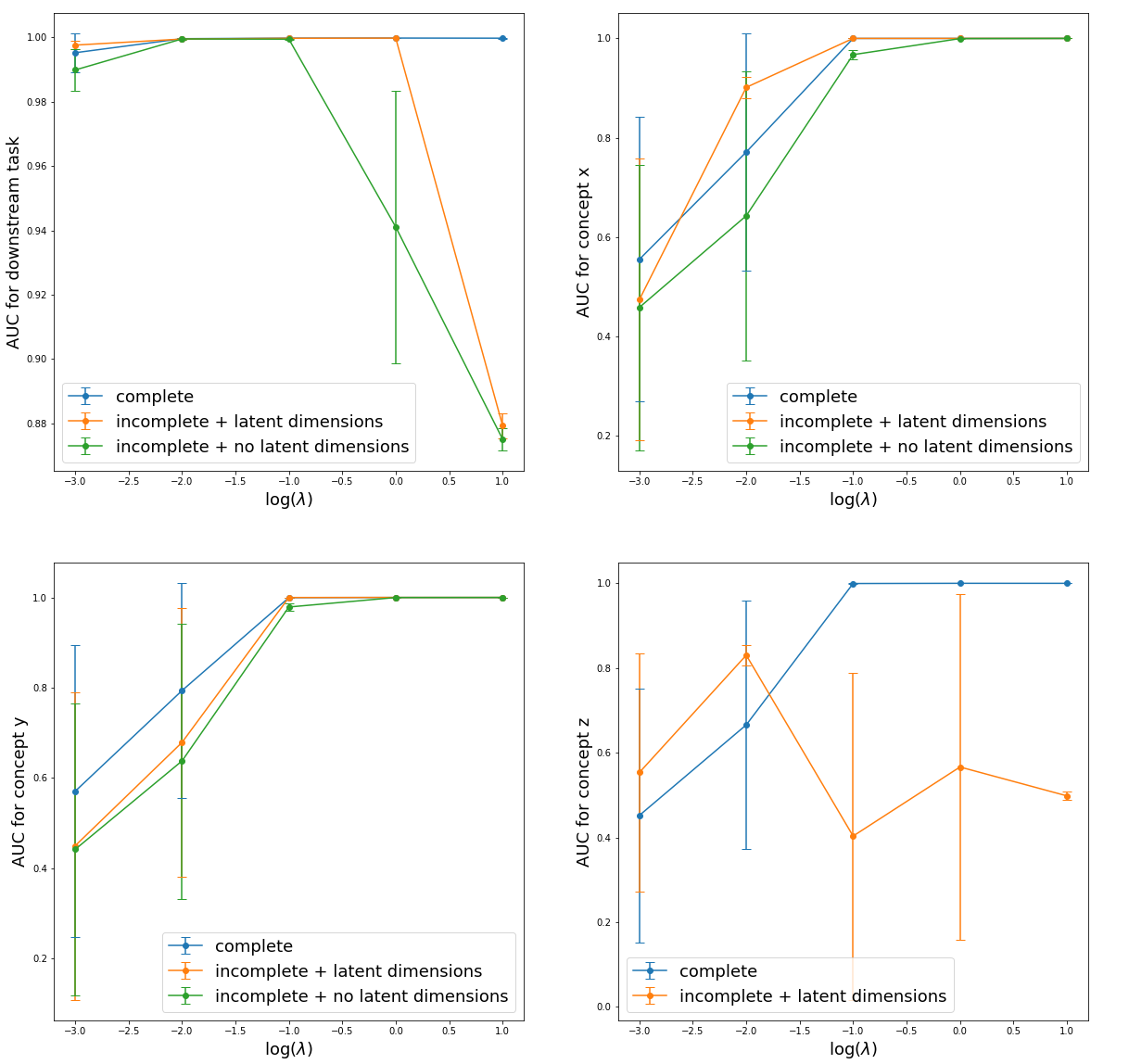}
    \caption{The three CBMs with  complete and incomplete concept sets are jointly trained with $\lambda =$ $0.001$, $0.01$, $0.1$, $1.0$, and $10.0$. $\lambda=0.1$ generally provides the best trade-off between downstream task and concept prediction AUC. The top left plot shows the AUC of the downstream task against $\log(\lambda)$.  The remaining plots display the AUC of predicting concepts $x_+$, $y_+$, and $z_+$ from the bottleneck dimension with which they are supposed to be aligned. The CBM with an incomplete concept set and one latent bottleneck dimension does not align $z_+$ with the latent bottleneck dimension during training; its AUC does not stay above $0.5$ across random restarts regardless of the value of $\lambda$.}
    \label{fig:incompleteness_cbm_r2_plots}
\end{figure*}

The CBM results discussed in the previous paragraph hint at potential impurities in the bottleneck dimensions. To evaluate concept purity, we compute AUC scores between the bottleneck dimension outputs and each of the ground truth concepts. For all the models (even when the full concept set is used during training), we observe impurities in all the bottleneck dimensions; although the aligned bottleneck dimensions have the highest AUC scores for the concepts with which they were aligned, they also have AUC scores that are larger than 0.6 for all the concepts with which they were not aligned (Tables  \ref{auc-complete}, \ref{auc-incomplete}, and \ref{auc-latent}). If the bottleneck dimensions were pure, these AUC scores should have been around 0.5. Having an incomplete concept set exacerbates the leakage problem as the model is forced to relay a lot more information about the missing concept through the bottleneck dimensions; the larger AUC scores of about 0.75 for the $z_+$ concept in Table \ref{auc-incomplete} are indicative of this problem.
Adding latent bottleneck dimensions in an attempt to capture the missing concept without compromising purity does not resolve the issue either. We notice that the latent bottleneck dimension does not consistently align with the missing concept across random trials as suggested by the AUC score of $0.4 \pm 0.4$. Furthermore, the latent dimension had the potential to leak information about the $x_+$ and $y_+$ concepts in some of the trials thus further compromising the interpretability of the model.

\begin{table}[t]
\caption{AUC scores between the concept values $x_{+}, y_{+}, z_{+}$ and bottleneck dimensions from the CBM with three bottleneck nodes trained with $\lambda=0.1$, aligning dimension 1 with $x_{+}$, dimension 2 with $y_{+}$, and dimension 3 with $z_{+}$}
\label{auc-complete}
\vskip 0.15in
\begin{center}
\begin{small}
\begin{sc}
\begin{tabular}{lcccr}
\toprule
\multicolumn{1}{}{} & \multicolumn{3}{c}{Bottleneck Dimension} \\
 & 1 & 2 & 3 \\
\midrule
$x_{+}$   & $0.9999 \pm 0.0001$ & $0.65 \pm 0.03$ & $0.63 \pm 0.03$  \\
$y_{+}$  & $0.65 \pm 0.02$  & $0.9995 \pm 0.0001$ & $0.67 \pm 0.04$  \\
$z_{+}$     & $0.64 \pm 0.02$ & $0.66 \pm 0.02$ & $0.9995 \pm 0.001$ \\
\bottomrule
\end{tabular}
\end{sc}
\end{small}
\end{center}
\vskip -0.1in
\end{table}

\begin{table}[t]
\caption{AUC scores between the concept values $x_{+}, y_{+}, z_{+}$ and bottleneck dimensions from the CBM with two bottleneck nodes trained with $\lambda=0.1$ and an incomplete concept set, aligning dimension 1 with $x_{+}$ and dimension 2 with $y_{+}$}
\label{auc-incomplete}
\vskip 0.15in
\begin{center}
\begin{small}
\begin{sc}
\begin{tabular}{lcccr}
\toprule
\multicolumn{1}{}{} & \multicolumn{2}{c}{Bottleneck Dimension} \\
 & 1 & 2  \\
\midrule
$x_{+}$   & $0.97 \pm 0.01$ & $0.64 \pm 0.01$   \\
$y_{+}$  & $0.66 \pm 0.01$ & $0.98 \pm 0.01$  \\
$z_{+}$     & $0.75 \pm 0.01$ & $0.74 \pm 0.01$ \\
\bottomrule
\end{tabular}
\end{sc}
\end{small}
\end{center}
\vskip -0.1in
\end{table}

\begin{table}[t]
\caption{AUC scores between the concept values $x_{+}, y_{+}, z_{+}$ and bottleneck dimensions from the CBM with 3 bottleneck nodes trained with $\lambda=0.1$ and an incomplete concept set, aligning dimension 1 with $x_{+}$ and dimension 2 with $y_{+}$.}
\label{auc-latent}
\vskip 0.15in
\begin{center}
\begin{small}
\begin{sc}
\begin{tabular}{lcccr}
\toprule
\multicolumn{1}{}{} & \multicolumn{3}{c}{Bottleneck Dimension} \\
 & 1 & 2 & 3 \\
\midrule
$x_{+}$   & $0.99998 \pm 0.00001$ & $0.59 \pm 0.02$ & $0.5 \pm 0.2$  \\
$y_{+}$  & $0.60 \pm 0.01$ & $0.99973 \pm 0.00002$ & $0.5 \pm 0.3$  \\
$z_{+}$     & $0.68 \pm 0.02$ & $0.693 \pm 0.005$ & $0.4 \pm 0.4$ \\
\bottomrule
\end{tabular}
\end{sc}
\end{small}
\end{center}
\vskip -0.1in
\end{table}

\section{Demonstration 4: Information Leakage in Concept Whitening Models}
\label{sec:demo3}

Since the CW model is primarily built for image recognition, we consider the task of classifying whether digits in MNIST are less than 4. We create the following three binary concepts: `is the number 1' ($c_1$), `is the number 2' ($c_2$), and `is the number 3' ($c_3$). We limit the dataset to images of numbers 1 to 6 in order to get a balanced classification problem. 

We set up a standard CNN with five layers (8, 16, 32, 16, and 8 filters each). All layers use a 3 by 3 kernel and a stride of 1. We apply batch normalization after each convolutional layer. 2 by 2 max pooling layers with strides of 2 are placed after the second, fourth, and fifth layers' outputs. Global average pooling is applied to the last layer's output and the flattened result is passed to a linear layer with a single node to make a prediction. The model is trained with an Adam optimizer and a learning rate of 0.001. We also set up a CW model by replacing the last batch normalization layer with the CW layer provided by \cite{chen2020concept} in the code accompanying their paper. We use the CW model to align the latent dimensions with concepts $c_1$ and $c_2$ and leave the third concept out of training. To confirm the CW model has finished aligning the latent dimensions to our concept examples, we use three quantitative checks that were reported in \cite{chen2020concept}: 1) reduced correlation between latent dimensions 2) an increase in AUC scores of the latent dimensions and 3) reduced inter-concept cosine similarity. 

We observe that correlation reduction is achieved early on during training; however, the model tends to struggle with achieving high AUC scores. More specifically, most trials tend to achieve a high AUC score (above 0.8) for one of the concepts without improving the AUC score for the latent dimension aligned with the other concept. We report results from a trial that was able to achieve high AUC scores of above 0.9 for both concepts to show that high AUC scores alone cannot be indicative of concept purity.

\begin{figure}
    \centering
    \includegraphics[width=\columnwidth]{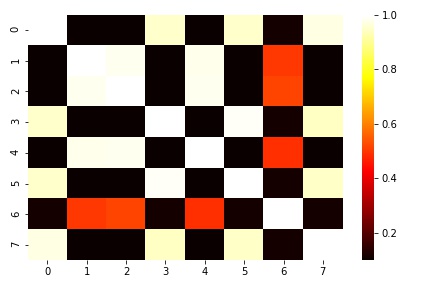}
    \includegraphics[width=\columnwidth]{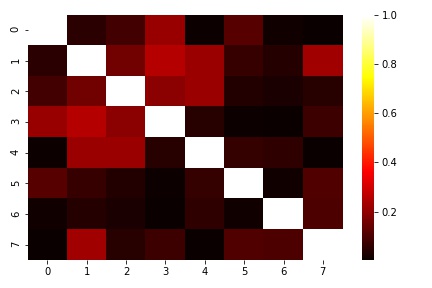}
    \caption{\textbf{Top:} Correlation matrix between the 8 dimensions outputted by the last batch normalization layer in the standard CNN. \textbf{Bottom:} Correlation matrix between the 8 dimensions outputted by the CW layer in the CW model. We observe less correlation between the CW layer dimensions compared to the standard CNN model.}
    \label{fig:correlation_plots}
\end{figure}

\begin{figure}
    \centering
    \includegraphics[width=\columnwidth]{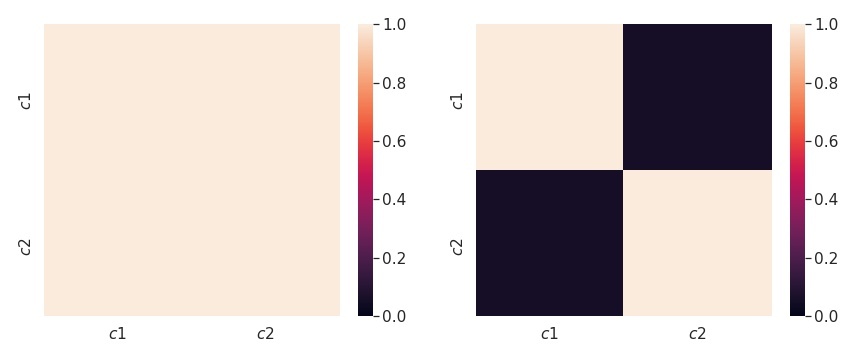}
    \caption{Normalized cosine similarities for examples of the same concepts (diagonal) and different concepts (off-diagonal) for the standard CNN model's last batch normalization layer output (left) vs. the CW model's output (right).}
    \label{fig:cos_similarity}
\end{figure}

Although all three concept purity checks that were reported in \cite{chen2020concept} are satisfied in this trial, we propose a different purity check that highlights impurities in the CW model's latent dimensions. Using the AUC score to evaluate purity may not provide a complete picture of how much information leakage there is between concepts in the model, since these scores are calculated using a single activation value from each of the latent dimensions (ie CNN filter outputs), while the rest of the CW model, has access to the full activation map of the latent dimension and is free to use any information from it. As a result, the AUC score may suggest the concept representation is pure, while the predictions based on the concept representation may contain information leakage between the concepts.

We demonstrate that this can occur by creating a purity check neural network with two convolutional layers (16 and 32 filters each) followed by a linear output layer. We provide the model with one of the output dimensions of the CW layer as inputs and train the model to predict each of the three concepts. We observe that the purity check model can predict any of the three concepts from any of the latent dimensions (aligned or not aligned with a concept) with over $95\%$ accuracy. This is despite the high AUC scores achieved by whitening and alignment. As a result, unless the rest of our model only has access to the same activation values used to compute AUC, we cannot be sure that the rest of the model is interpreting the latent dimensions as intended (ie aligned with a specific concept) thus potentially rendering meaningless, any interpretation of feature importance of the downstream task model. For instance, in our experiment, we examine the final linear layer's weights which are applied to the global average pooling values for each of the CW layer's dimensions and observe that the first weight is 0.874 while the second weight is 0.056. The first and second CW dimensions were aligned with the concepts $c_1$ and $c_2$ which are equally important for predicting the downstream task labels defined as $c_1 + c_2 + c_3 > 1$. However, if we were to interpret the CW model's weights, we would conclude that $c_1$ is more important than $c_2$.

\section{Demonstration 5: Concept Refinement}
\label{sec:demo4}
\begin{figure}[h!]
    \centering
    \includegraphics[scale=0.5]{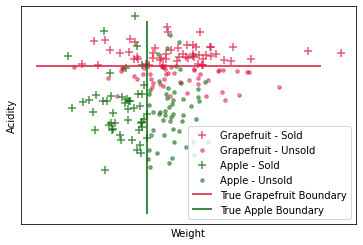}
    \caption{Fruit sales by weight and acidity}
    \label{fig:hierarchical_concepts}
\end{figure}

Consider a synthetic task where we model which type of fruit would sell based on weight and acidity. We identify two concepts potentially relevant to the prediction task: `is grapefruit'' and ``is apple''. Figure \ref{fig:hierarchical_concepts} shows our synthetic dataset, where the true decision boundaries depict the fact that people like to buy acidic grapefruit and smaller apples. If we were to model this problem using just acidity and weight and not the type of fruit, our model would perform well but would not provide insight on what distinguishes small apples that do sell from small grapefruit that do not. If we were to instead use the concepts of ``is grapefruit'' and ``is apple'' to predict whether or not our fruits sell, we would find that these concepts are not predictive at all -- approximately 50\% of grapefruit and apples sell. 

However, we see from Figure \ref{fig:hierarchical_concepts} that though these concepts are not predictive on their own, they are related to the problem. If we were to replace the concepts of ``is grapefruit'' and ``is apple'' with ``is acidic grapefruit'' and ``is small apple'', i.e. by splitting our original concepts into sub-concepts, our concept set would be perfectly predictive. But what's the difference between this and using all three of weight, acidity and fruit as features? Given proper thresholds, weight and acidity can be reduced to ``is heavy'' and ``is acidic''. Thus, by finding a good refinement of our original concept set, we can nonetheless reduce the input space to a small number of regions that are is predictive, and when the dimensional of the data is high, concept refinement can still yield interpretable but compressed representations of the data.

\end{document}